\documentclass[a4paper,fleqn]{cas-dc}

\usepackage[authoryear,longnamesfirst]{natbib}
\usepackage{ulem}
\def\tsc#1{\csdef{#1}{\textsc{\lowercase{#1}}\xspace}}
\tsc{WGM}
\tsc{QE}
\begin{document}
\let\WriteBookmarks\relax
\def\floatpagepagefraction{1}
\def\textpagefraction{.001}

\shorttitle{Multi-view self-supervised learning and two-stage pre-training}    

\shortauthors{J.W., X.Y.}  

\title [mode = title]{Thyroid ultrasound diagnosis improvement via multi-view self-supervised learning and two-stage pre-training}  

% \tnotemark[<tnote number>] 



% first author
\author[1]{Jian Wang}

% co-author
\author[2,3,4]{Xin Yang}
\author[5]{Xiaohong Jia}
\author[2,3,4]{Wufeng Xue}

\author[2,3,4]{Rusi Chen}
\author[2,3,4]{Yanlin Chen}
\author[2,3,4]{Xiliang Zhu}
\author[2,3,4]{Lian Liu}
\author[6]{Yan Cao}


\author[5]{Jianqiao Zhou}
\cormark[1]
\ead{zhousu30@126.com}

\author[2,3,4]{Dong Ni}
\cormark[1]
\ead{nidong@szu.edu.cn}

\author[1,7]{Ning Gu}
\cormark[1]
\ead{guning@nju.edu.cn}

% % Footnote of the first author
% \fnmark[<footnote mark no>]

% \ead{guning@nju.edu.cn}

% \ead[url]{<URL>}

\credit{<Credit authorship details>}

\affiliation[1]{organization={Key Laboratory for Bio-Electromagnetic Environment and Advanced Medical Theranostics, School of Biomedical Engineering and Informatics,Nanjing Medical University},
            city={Nanjing},
            postcode={211166}, 
            country={China}}

\affiliation[2]{organization={National-Regional Key Technology Engineering Laboratory for Medical Ultrasound, School of Biomedical Engineering, Health Science Center, Shenzhen University},
            city={Shenzhen},
            postcode={518073}, 
            country={China}}

\affiliation[3]{organization={Marshall Laboratory of Biomedical Engineering, Shenzhen University},
            city={Shenzhen},
            postcode={518073}, 
            country={China}}

\affiliation[4]{organization={Medical UltraSound Image Computing (MUSIC) Lab, Shenzhen University},
            city={Shenzhen},
            postcode={518073}, 
            country={China}}

\affiliation[5]{organization={Department of Ultrasound, Ruijin Hospital, Shanghai Jiaotong University School of Medicine},
            city={Shanghai},
            postcode={200025}, 
            country={China}}

\affiliation[6]{organization={Shenzhen RayShape Medical Technology Co., Ltd},
            city={Shenzhen},
            postcode={518051}, 
            country={China}}
            
\affiliation[7]{organization={Cardiovascular Disease Research Center, Nanjing Drum Tower Hospital, Affiliated Hospital of Medical School, Medical School, Nanjing University},
            city={Nanjing},
            postcode={210093}, 
            country={China}}

% \author[1]{Jian Wang}%[options]

% Footnote of the second author

% Email id of the second author
% \ead{}



% \affiliation[<aff no>]{organization={},
% %          citysep={}, % Uncomment if no comma needed between city and postcode

\cortext[1]{Corresponding author}



\begin{abstract}
Thyroid nodule classification and segmentation in ultrasound images are crucial for computer-aided diagnosis; however, they face limitations owing to insufficient labeled data. In this study, we proposed a multi-view contrastive self-supervised method to improve thyroid nodule classification and segmentation performance with limited manual labels. Our method aligns the transverse and longitudinal views of the same nodule, thereby enabling the model to focus more on the nodule area. We designed an adaptive loss function that eliminates the limitations of the paired data. Additionally, we adopted a two-stage pre-training to exploit the pre-training on ImageNet and thyroid ultrasound images. Extensive experiments were conducted on a large-scale dataset collected from multiple centers. The results showed that the proposed method significantly improves nodule classification and segmentation performance with limited manual labels and outperforms state-of-the-art self-supervised methods. The two-stage pre-training also significantly exceeded ImageNet pre-training.
\end{abstract}


% \begin{highlights}
% \item We proposed a multi-view SSL method that is not constrained by paired views.
% \item We adopted a two-stage pre-training strategy on thyroid ultrasound images.
% \item Extensive experiments were conducted on a large thyroid ultrasound image dataset.
% \end{highlights}

% Keywords
\begin{keywords}
 Thyroid ultrasound image\sep Self-supervised learning\sep Multi-view learning\sep 
 Two-stage pre-training\sep Nodule classification\sep Nodule segmentation\sep
\end{keywords}

\maketitle

% Main text
\section{Introduction}\label{}

Thyroid cancer is among the most common cancers worldwide \citep{sung2021global}. Early detection enables timely intervention and avoids overdiagnosis \citep{bethesda2018seer}. Ultrasound is the primary imaging tool for thyroid diagnosis because it is real-time, non-invasive, and low-cost \citep{smith2012use}. However, accurate interpretation of ultrasound images requires experienced physicians, and inexperienced physicians may misdiagnose. Several computer-aided diagnostic methods have been proposed to address this issue, most of which are based on deep learning. For example, \citet{deng2022automatic} proposed a multi-task network to determine Thyroid Imaging Reporting and Data System grade for identifying the benignity and malignancy of thyroid nodules. \citet{sun2023classification} proposed a contrast-learning-based thyroid nodule classification model to improve the accuracy of diagnosis. \citet{kang2022thyroid} proposed intra- and inter-task consistent learning to enforce the network to learn consistent predictions for nodule classification and segmentation. \citet{gong2023thyroid} designed a multi-task learning framework to accurately segment the thyroid nodule. For more related work, please refer to \citet{chen2020review,sharifi2021deep}.

\begin{figure}[!t]
\centering
\includegraphics[scale=0.34]{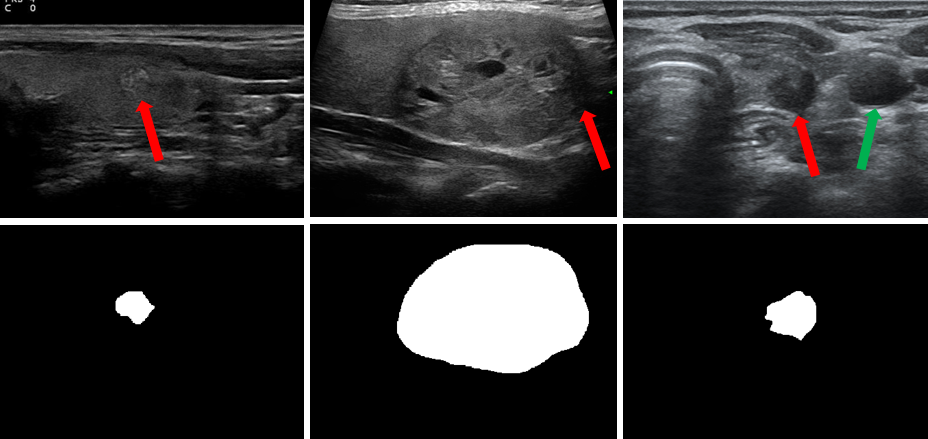}
\caption{The upper row is the thyroid ultrasound images of different patients, and the lower row is the corresponding nodule masks. Red arrows indicate nodules and green arrows indicate carotid vessels.}
\label{fig:anno}
\end{figure}

Although these methods have the potential to solve the aforementioned clinical problem, they require large amounts of data with manual annotations. The fine-needle aspiration (FNA) biopsy is the gold standard for distinguishing benign and malignant nodules. However, only a small fraction of patients undergo an FNA biopsy. For nodule segmentation, the gold standard is obtained by manually delineating the pixel-level mask. As shown in Fig.\ref{fig:anno}, the borders of the nodules in the image are often blurred and incomplete and the nodules may resemble carotid vessels. Therefore, annotating the mask requires a comprehension of the thyroid anatomy, recognition of relevant features from the ultrasound image, and exclusion of interference from similar tissues. Annotating ultrasound images is time-consuming and requires costly expertise that is not easily accessible. Most labeled thyroid image datasets are thus small and lack manual labels. This limits the development of deep learning techniques in thyroid image analysis.

Recently, self-supervised learning (SSL) has been applied to various medical images to address the problem of insufficient labeled data. Its core objective is to provide good initialization for the target task by pre-training the model without using artificial labels. For example, \citet{zhou2021models} used image restoration to pre-train models on 2D chest CT slices, chest X-ray images, and 3D chest volumes. \citet{zhou2020comparing} used contrastive learning to pre-train the model on chest X-ray images. \citet{zhu2020rubik} used 3D jigsaw puzzles to pre-train the model on brain CT and MRI volumes. \citet{punn2022bt} pre-trained models on breast ultrasound images and dermoscopy images via redundancy reduction. \citet{basu2022unsupervised} combined contrastive learning and hard negative mining to pre-train models on gallbladder ultrasound videos. For more information about the medical imaging applications of SSL, please refer to \citet{shurrab2022self}.

\begin{figure}[!t]
  \centering
  \includegraphics[scale=0.29]{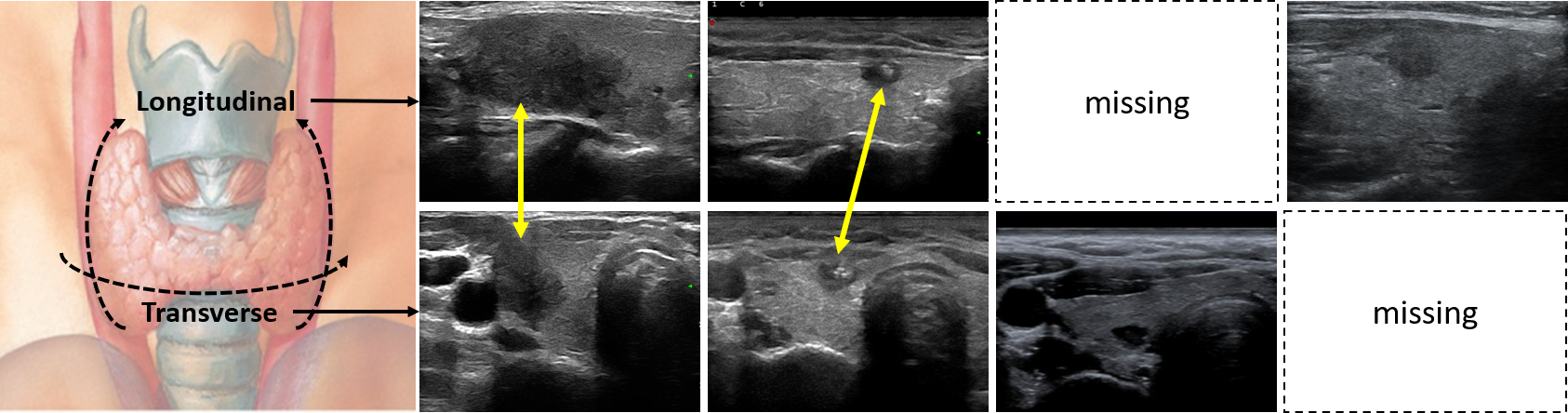}
  \caption{The color image on the left is a schematic of the thyroid, and thyroid ultrasound images from four patients are on the right. The upper row is the longitudinal views, and the lower row is the corresponding transverse views. The yellow arrows point to the same nodule.}
  \label{fig:views}
  \end{figure}

We investigated a large number of SSL-related research in the field of medical image analysis from the literature and found that the existing methods have some limitations. \textit{First}, most methods were designed for single-view tasks, which does not consider the multi-view nature of the thyroid. Physicians usually scan the thyroid gland of a patient horizontally and vertically to conduct a complementary examination, resulting in transverse and longitudinal views. It is also possible that one of the two views is missing. As shown in Fig.\ref{fig:views}, the lower row shows thyroid transverse views from four patients, and the upper row shows the corresponding longitudinal views. Two views of the same nodule display relevant and complementary information, such as nodule shape and echo pattern. Moreover, two views of the same nodule should belong to the same category, either benign or malignant. Intuitively, exploiting such multi-view consistency can enhance SSL, thereby improving the target task performance. \textit{Second}, the SSL methods designed for multimodal data can theoretically be used for multi-view thyroid images \citep{hervella2020self, hervella2021self, fedorov2021self, fedorov2021self2, li2020self, taleb2017self, xiang2022self}, but they cannot handle missing views well. This is because they all require the paired data, and unpaired data cannot be used. \textit{Third}, the combination of SSL-based pre-training on medical images and pre-training on natural images may benefit the target task but was ignored by existing self-supervised studies. Most self-supervised studies have demonstrated that self-supervised pre-training on medical images can outperform supervised pre-training on natural images for target tasks. However, one can easily collect numerous natural images from the Internet, but it is difficult to obtain numerous medical images from hospitals, even without labels. 

In this study, we attempt to improve the nodule classification and segmentation performance of thyroid ultrasound images with limited manual labels using a novel SSL method. The main contributions of our study are as follows: (1) We proposed a multi-view SSL method for thyroid ultrasound image analysis. To address the issue of missing views, we designed an adaptive loss function that allowed the model to utilize unpaired views. (2) We adopted a two-stage pre-training strategy to combine ImageNet pre-training and self-supervised pre-training on thyroid ultrasound images, which exploits the benefits of both. (3) Extensive experiments were conducted on a large-scale thyroid ultrasound image dataset collected from multiple centers and devices. The experimental results showed that the proposed method significantly improves nodule classification and segmentation performance with limited manual labels, demonstrating its effectiveness.

\section{Related work}

In this section, we briefly introduce three fields related to our research: self-supervised learning, multi-view learning, and two-stage pre-training.

\subsection{Self-supervised learning}

SSL can be classified into three main categories: predictive, generative, and contrastive \citep{shurrab2022self}. \textit{Predictive} methods usually first transform the image and then use models to predict these transformations. Examples include the jigsaw puzzle \citep{noroozi2016unsupervised}, relative-position prediction \citep{doersch2015unsupervised}, and rotation prediction \citep{gidaris2018unsupervised}. However, these methods must be carefully designed to prevent the network from taking shortcuts to learn meaningless representations for the target tasks. \textit{Generative} methods usually use an encoder-decoder structure for image reconstruction, such as image inpainting \citep{pathak2016context}, image context restoration \citep{chen2019self}, models genesis \citep{zhou2021models}. Such methods often provide limited improvements in the target performance. Recently, breakthroughs have been achieved in visual transformers (ViTs) based on generative methods \citep{he2022masked}. Partial patches of the image were masked, and visible patches were sent to a ViT-based network for patch reconstruction. Although these methods produce more generalizable visual features, they are specifically designed for ViT architecture and are computationally intensive. Additionally, without further fine-tuning, pre-trained features may not perform favorably in some scenarios. \textit{Contrastive} methods aim to minimize the feature distances of positive pairs while maximizing the feature distances between negative pairs. Positive pairs usually refer to different augmented versions of the same image, and negative pairs usually refer to different images. The representative methods include SimCLR \citep{chen2020simple} and MoCo \citep{he2020momentum, chen2020improved}. These methods have been shown to outperform supervised ImageNet pre-training. In this study, we extend the common contrastive approach by considering different thyroid ultrasound views of the same nodule as positive pairs.

\subsection{Multi-view learning}

Multi-view learning is a scenario where representations are learned by correlating information from multiple views of data to improve learning performance \citep{li2018survey}. Multi-view \textit{supervised learning is an active research area \citep{wang2020auto, wu2022multi, shah2023multi, kim2022emotion}}. For example, \citet{wang2020auto} proposed a multimodal fusion network that fuses B-mode, Doppler, shear wave, and strain wave ultrasound images to predict benign and malignant breast nodules. In multi-view \textit{self-supervised learning}, \citet{hervella2020self,hervella2021self} proposed a generative framework that reconstructs gray angiography from the corresponding colorful retinography. \citet{fedorov2021self, fedorov2021self2} used a contrastive framework on multimodal MRI images to maximize the mutual information. \citet{xiang2022self} proposed a self-supervised multi-modal fusion network on multimodal thyroid images. \citet{hassani2020contrastive} proposed a contrastive framework on graphs. \citet{roy2021self} proposed a contrastive framework on human facial images. However, these SSL methods require paired views. To address this problem, \citet{li2020self} first used CycleGAN \citep{zhu2017unpaired} to synthesize missing modalities from other modalities. They then pre-trained the model using a contrastive self-supervised framework that aligned multimodal features. Similarly, \citet{taleb2017self} also used CycleGAN to synthesize the missing modality and designed multimodal jigsaw puzzles on multimodal MRI images. However, the quality of synthetic data is difficult to evaluate and may be detrimental to pre-training. In contrast, the proposed method does not require paired data. Additionally, our method is evaluated on image classification, segmentation, and multi-view classification, which has not been verified in other studies.

\subsection{Two-stage pre-training}

Given the significant differences between natural and medical images, transfer learning from natural to medical images may be suboptimal \citep{raghu2019transfusion}. Therefore, some studies have proposed a two-stage pre-training for target tasks. For example, \citet{liu2021covid} first initialized the partial layers of their proposed network with weights pre-trained on ImageNet and continued to pre-train the model on numerous labeled pulmonary nodule CT images before fine-tuning it for COVID-19 lung infection segmentation. Similarly, \citet{meng2022tl} used two-stage pre-training to initialize the model and used it for COVID-19 image classification. \citet{zhang2022two} built a large dataset from natural images similar to tongue manifestation images and trained the pre-trained model on it before fine-tuning it on real clinical tongue manifestation images for target tasks. Although these studies demonstrated the effectiveness of two-stage pre-training, it is challenging to collect a large amount of labeled data for supervised learning in the second stage. A natural idea is to replace supervised pre-training with self-supervised pre-training in the second stage, which exploits a large amount of unlabeled data.  However, only a few studies \citep{azizi2021big,verma2022can} have attempted this strategy, and a detailed analysis is lacking. In this study, we carefully explore two-stage pre-training in two ways: supervised to self-supervised and self-supervised to self-supervised.

\section{Method}

In this section, we first introduce the proposed SSL framework and then propose our adaptive loss. Finally, we introduce the two-stage pre-training and implementation details.

\begin{figure*}[!t]
  \centering
  \includegraphics[scale=0.49]{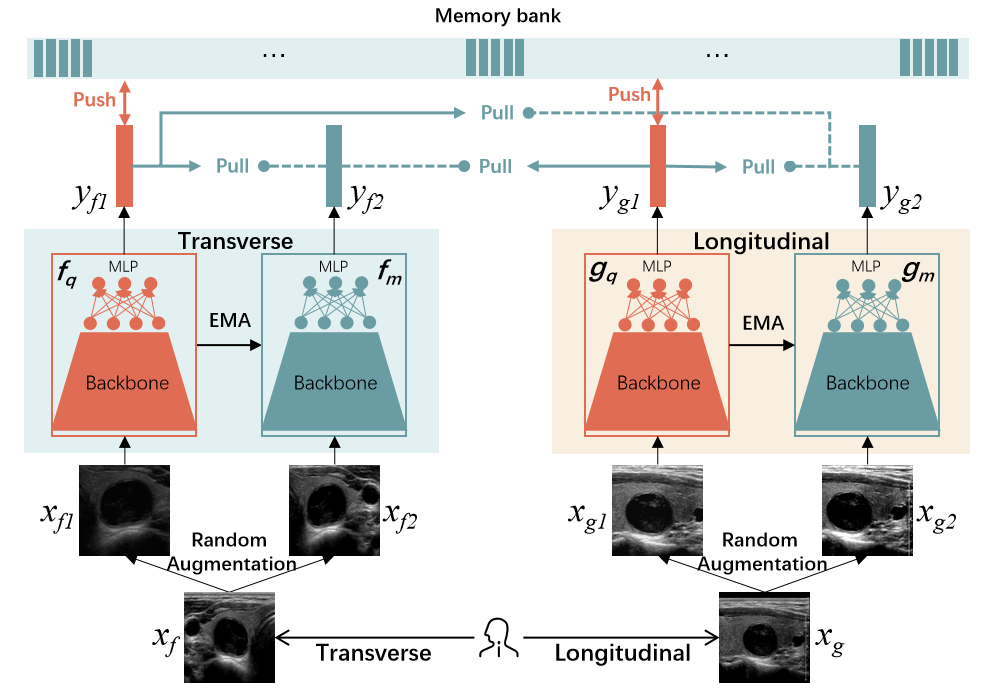}
  \caption{Our framework adopts independent query and momentum encoders for each view, and the two views share the same memory bank.}
  \label{fig:framework}
  \end{figure*}

\subsection{Proposed framework}

Common contrastive learning methods regard different data-augmented versions of the same image as positive pairs, and different images as negative pairs. They reduce the feature distance between positive pairs and increase the feature distance between negative pairs. This allows the model to be trained without manual labels and to achieve a target task performance that meets or exceeds the level of supervised ImageNet pre-training \citep{chen2020simple, chen2020improved}. However, such methods may lead to multi-view images of the same nodule being assigned to different categories. In thyroid ultrasound scanning, the transverse and longitudinal views of the same nodule are visually related. As shown in Fig.\ref{fig:views}, two views of the same nodule are related and complementary in terms of nodule shape and echo pattern and are differentiated in views of different nodules. Furthermore, two views of the same nodule share the same benign and malignant categories. Therefore, common contrastive learning methods may not be suitable for multi-view thyroid ultrasound images.

To solve this problem, we propose a multi-view contrastive self-supervised framework. Fig.\ref{fig:framework} shows this framework. Inspired by MoCo v2 \citep{chen2020improved}, we adopt two sets of encoders for the transverse and longitudinal views, where each set of encoders consists of a query and a momentum encoder. For convenience, we denote the transverse and longitudinal encoders as $\textit{f}(\cdot)$ and $\textit{g}(\cdot)$, and use the subscripts $\textit{q}$ and $\textit{m}$ to represent the query and momentum encoder, respectively. The four encoders ($\textit{f}_{q}(\cdot)$, $\textit{f}_{m}(\cdot)$, $\textit{g}_{q}(\cdot)$, and $\textit{g}_{m}(\cdot)$) share the same network architecture and initial weights. Each encoder consists of a backbone and a projection head. The backbone and projection are convolutional neural networks (CNNs) and multilayer perception (MLP), respectively. The backbone is used to extract features from the images in the input space, and the projection head is used to project the extracted features into the latent space. After pre-training, only the backbone of the query encoder is used for the target tasks, whereas the projection head is discarded. Additionally, we adopt a memory bank that can store \textit{K} vectors. This memory bank stores the vectors projected onto the latent space and works as a queue that follows the first-in-first-out principle. 

Given the paired views of the same patient from a batch $\textbf{\textit{X}}$, we denote the transverse view as $\textbf{\textit{x}}_f$ and the longitudinal view as $\textbf{\textit{x}}_g$. We employ the random data augmentation on $\textbf{\textit{x}}_f$ and $\textbf{\textit{x}}_g$ to generate two augmented versions ($\textbf{\textit{x}}_{f1}$ and $\textbf{\textit{x}}_{f2}$, $\textbf{\textit{x}}_{g1}$ and $\textbf{\textit{x}}_{g2}$). The augmented transverse views, $\textbf{\textit{x}}_{f1}$ and $\textbf{\textit{x}}_{f2}$ are fed into $\textit{f}_{q}(\cdot)$ and $\textit{f}_{m}(\cdot)$, respectively. Similarly, the augmented longitudinal views, $\textbf{\textit{x}}_{g1}$ and $\textbf{\textit{x}}_{g2}$ are fed into $\textit{g}_{q}(\cdot)$ and $\textit{g}_{m}(\cdot)$, respectively. After feature extraction and projection, the corresponding vectors, $\textbf{\textit{y}}_{f1}$, $\textbf{\textit{y}}_{f2}$, $\textbf{\textit{y}}_{g1}$, and $\textbf{\textit{y}}_{g2}$ are obtained, where $\textbf{\textit{y}}_{f1}$ and $\textbf{\textit{y}}_{f2}$, $\textbf{\textit{y}}_{g1}$ and $\textbf{\textit{y}}_{g2}$ are considered as two positive pairs since they are from the same image ($\textbf{\textit{x}}_f$ and $\textbf{\textit{x}}_g$). The loss function is described in the following subsection. After computing the loss, the weights of query encoders ( $\textit{f}_{q}(\cdot)$ and $\textit{g}_{q}(\cdot)$ ) are updated through back-propagation, and the weights of momentum encoders ( $\textit{f}_{m}(\cdot)$ and $\textit{g}_{m}(\cdot)$ ) are updated by exponential moving average (EMA):
\begin{equation}
  \theta_m^t \gets \alpha\cdot\theta_m^{t-1}+(1-\alpha)\cdot\theta_q^t,
  \end{equation}
where $\theta_q$ and $\theta_m$ denote the weights of the query and momentum encoders, respectively. The superscript \textit{t} denotes the training step, and $\textit{$\alpha$} \in \left [0,1  \right ]$ is the momentum coefficient that controls the speed of the weight update. After weight updating, $y_{f2}$ and $y_{g2}$ are sent to the memory bank as new vectors, and the oldest vectors in the memory bank are dequeued. In our framework, the query encoders, $\textit{f}_{q}(\cdot)$ and $\textit{g}_{q}(\cdot)$ share the same weights, whereas the momentum encoders, $\textit{f}_{m}(\cdot)$ and $\textit{g}_{m}(\cdot)$ share the same weights. We also tried separate weights for the transverse and longitudinal encoders but found it better to use the weight-sharing mechanism.

\subsection{Adaptive loss}

Our loss function comprises two parts: single-view and cross-view contrastive losses. The single-view contrastive loss reduces the feature distance between different augmented versions of the same image, enabling the encoder to learn transformation-invariant features. It can be written as:
\begin{equation}
  \begin{split}
  L_{ff}=C(y_{f1},y_{f2}), \\
  L_{gg}=C(y_{g1},y_{g2}),
\end{split}
\end{equation}
where $\textit{C}(\cdot)$ is the InfoNCE loss \citep{oord2018representation}, which can be expressed as:
\begin{equation}
  \begin{aligned}
C(q,k)=
-\log \frac{exp(sim(q,k)/\tau)}
{exp(sim(q,k)/\tau)+{\textstyle \sum_{i=1}^{N}}exp(sim(q,t_i)/\tau)},
\label{con:infonce}
  \end{aligned}
\end{equation}
where $\tau$ is a temperature parameter and $\textit{sim}(\cdot)$ is the operator for similarity measurement; here, it is set as cosine similarity. The vector \textit{q} and vector \textit{k} are a positive pair, whereas vector \textit{q} and vector $t_i$ ($\textit{i} \in \left \{ 1,2,...,N \right \}$) are a negative pair. In this study, the vector ($\textit{y}_{f1}$ and $\textit{y}_{g1}$) generated by query encoders ($\textit{f}_{q}(\cdot)$ and $\textit{g}_{q}(\cdot)$) and each vector stored in the memory bank are considered negative pairs because they are from different images or from the same image but at different training steps. Thus, there are \textit{K} negative pairs and a positive pair, which forms the log loss of a (\textit{K}+1)-way softmax-based classifier.

The cross-view contrastive loss reduces the feature distance between the transverse and longitudinal views of the same nodule, which enables the encoder to learn view-invariant features. It can be expressed as:
\begin{equation}
  \begin{split}
  L_{fg}=C(y_{f1},y_{g2}),\\  
  L_{gf}=C(y_{g1},y_{f2}).
\end{split}
\end{equation}
A naive idea is to combine these losses directly in the form of a summation as follows:
\begin{equation}
  L_{pair}=L_{ff}+L_{gg}+L_{fg}+L_{gf}.
  \label{con:naive}
\end{equation}
Obviously, this loss function is only used for paired views, and those unpaired views cannot be utilized. Unfortunately, it is possible to happen that one of the two views is missing in clinical practice. To solve this problem, \citet{li2020self,taleb2017self} used CycleGAN to synthesize the missing views before pre-training. However, image synthesis itself requires paired data, and synthetic images may be harmful to pre-training owing to poor synthesis quality. Therefore, we propose the following adaptive loss function:
\begin{equation}
  L_{adaptive}=a\cdot L_{ff}+b\cdot L_{gg}+ab\cdot\lambda (L_{fg}+L_{gf}),
  \label{con:adaptive}
\end{equation}
where $\textit{a},\textit{b} \in \left \{ 0,1 \right \}$ are two indicator functions that evaluate zero if the corresponding view is missing, and $\lambda$ is a coefficient that balances the single-view and cross-view contrastive losses. If one of the two views is missing, $L_{adaptive}$ is equal to $L_{ff}$ or $L_{gg}$, which means that it adaptively becomes a single-view contrastive loss. This eliminates the limitation of requiring paired views. The final loss function is the average of the loss functions corresponding to each patient in the batch $\textbf{\textit{X}}$. In summary, this adaptive loss function has two advantages: (1) The encoder can learn both transformation-invariant and view-invariant features, which is superior to only learning transformation-invariant features. (2) The encoder can be optimized on both single-view data and multi-view data, which is flexible and does not require paired data.

\begin{figure}[!t]
  \centering
  \includegraphics[scale=0.36]{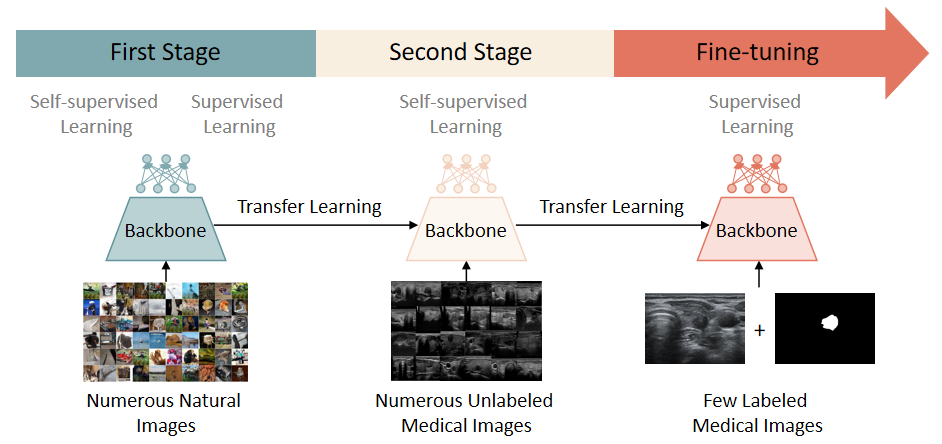}
  \caption{Two-stage pre-training. In the first stage, we train the model on ImageNet in a supervised and self-supervised learning manner. In the second stage, we first initialize the model with the learned weights from the first stage and train the model on unlabeled target medical images in a self-supervised manner. Finally, the model is fine-tuned for the target tasks.}
  \label{fig:two_stage}
  \end{figure}
\subsection{Two-stage pre-training}
For medical image analysis, self-supervised pre-training on medical images can alleviate the domain shift between ImageNet and medical datasets compared with ImageNet pre-training. However, it is also limited by the challenge of collecting numerous medical images, owing to patient privacy. In this study, instead of using one-stage pre-training, we divided the pre-training process into two stages, as shown in Fig.\ref{fig:two_stage}. In the first stage, we trained the model on ImageNet in a supervised and self-supervised manner. This enables the model to learn the features of natural images and provides good initialization for medical image analysis. However, this ability is susceptible to domain shift and may not be effective for certain medical images \citep{tajbakhsh2016convolutional}. In the second stage, we further trained the model on unlabeled medical images similar to the target images in a self-supervised manner. This enables the model to learn the features of medical images and enhances its ability to handle such medical image analysis tasks. The pre-trained weights of common network architectures on ImageNet are usually available in the deep learning community. We can skip the first stage and directly initialize our model using these weights in the second stage. After the two-stage pre-training, we fine-tuned our model for target tasks using a small number of labeled target images in a supervised learning manner.

\subsection{Implementation details}

We divided the implementation into four aspects: network architecture, data augmentation, hyperparameters, and optimization settings. For the network architecture, we used ResNet50 \citep{he2016deep} as the backbone and a two-layer fully connected layer as the projection head. The hidden and output dimensions of the projection head are 512 and 128, respectively. For data augmentation, we employed common techniques, including cropping, resizing, brightness adjustment, contrast adjustment, Gaussian blur, horizontal flip, and rotation. For the hyperparameters, the temperature parameter $\tau$ was set to 0.1, and the memory bank size $\textit{K}$ was set to 512. The momentum coefficient $\alpha$ was set to 0.99 and the loss coefficient $\lambda$ was set to 0.5. For the optimization settings, we used the following details: SGD optimizer, cosine learning rate decay scheduler, initial learning rate of 0.03, weight decay of 0.0001, batch size of 128 (64 patients per iteration), and 200 training epochs. The intensity normalization (i.e. subtract the mean and divide by the standard deviation) was performed. To demonstrate that the target tasks benefit from our method rather than from special tricks, we also implemented MoCo v2 using the above settings. For the two-stage pre-training, we modified the memory bank size $\textit{K}$ to 1024 and the initial learning rate to 0.01. We obtained the supervised ImageNet pre-trained weights from PyTorch official repository and self-supervised ImageNet pre-trained weights from MMSelfSup \cite{mmselfsup2021}. All experiments in this study were performed on a PyTorch platform using servers equipped with NVIDIA RTX A40 GPUs.

\begin{table*}[htpb]
  \caption{\label{comm} Summary of the compared methods. These are the current SOTA self-supervised methods in medical image analysis. The categories include generative, contrastive, and combinations of the two. The last column refers to the images used in the original papers. }
  \centering
  \begin{tabular}{p{4.5cm}p{2.0cm}p{1.0cm}p{3.5cm}p{4.0cm}}

  \toprule
  \textbf{Methods} & \textbf{Publication} & \textbf{Years} & \textbf{Category}& \textbf{Images for pre-training}\\

  \midrule
  MoCo v2 \citep{he2020momentum} &
  CVPR &
  2020 & 
  contrastive & 
  ImageNet images \\ 

  C2L \citep{zhou2020comparing}& 
  MICCAI &
  2020 & 
  contrastive & 
  Chest X-ray images \\

  \multirow{3}{*}{MG \citep{zhou2021models}}& 
  \multirow{3}{*}{MedIA} &
  \multirow{3}{*}{2020} &
  \multirow{3}{*}{generative} &
  2D Chest CT slices\newline
  3D Chest CT volumes\newline
  Chest X-ray images\\  

  PCRL \citep{zhou2021preservational}& 
  \multirow{2}{*}{ICCV}& 
  \multirow{2}{*}{2021}& 
  \multirow{2}{*}{contrastive \& generative}& 
  Chest X-ray images\newline
  3D Chest CT volumes\\

  CAiD \citep{taher2022caid}& 
  \multirow{2}{*}{MIDL}& 
  \multirow{2}{*}{2022}&
  \multirow{2}{*}{contrastive \& generative}&
  \multirow{2}{*}{Chest X-ray images}\\
  
  DiRA \citep{haghighi2022dira}& 
 \multirow{2}{*}{CVPR}& 
 \multirow{2}{*}{2022}&
 \multirow{2}{*}{contrastive \& generative}&
  Chest X-ray images\newline 
  3D Chest CT volumes \\

  SSFL \citep{li2020self}& 
  TMI &
  2020 & 
  contrastive & 
  Multimodal fundus images \\ 

  \bottomrule
  \end{tabular}
  \end{table*}

\section{Materials and Experiments}

In this section, we introduce our large-scale thyroid ultrasound dataset collected from multiple centers, compared methods, and target tasks.

\subsection{Dataset}

To evaluate the proposed method, we constructed a large-scale thyroid ultrasound dataset. Our dataset has the following characteristics: (1) It was collected from multiple centers consisting of more than 20 hospitals and sites. These centers are located in different regions of China, such as Shanghai, Chengdu in Sichuan, and Changzhou in Jiangsu, providing regional diversity and physician scanning diversity. (2) Images were generated using more than 30 types of ultrasound imaging equipment under different settings. The equipment primarily included the Esaote Mylab series, GE LOGIQ E9, Mindray Resona7 series, Philips EPIQ7, SIEMENS ACUSONS 2000, SAMSUNG RS80A, and TOSHIBA Aplio series, which provides diverse imaging equipment. (3) This dataset contains 5224 patients ranging in age from 9 to 82 years old, providing patient diversity. This diverse dataset ensured the reliability of our experimental results. 

Our dataset contains 2216 patients with benign nodules and 3008 patients with malignant nodules. All nodule labels were determined using FNA biopsy reports. For the ultrasound images with multiple nodules, the labels were annotated as malignant if one nodule was malignant. This is the largest multicenter thyroid ultrasound image dataset with pathological labels. Additionally, the nodule masks were annotated by experienced physicians. Only the largest nodule in the image was labeled. For preprocessing, we first used the largest connected component algorithm to obtain the largest connected region. Second, we cut the surrounding regions to extract it as the region of interest (ROI). Third, the extracted ROIs were cropped or padded and then resized to $256\times256$ pixels. Finally, we manually verified all images and corrected the errors.

Our dataset contains 9669 images from 5224 patients. It is divided into two groups. The first group includes 779 patients with a single transverse or longitudinal view. The second group contains 4445 patients with both transverse and longitudinal views. Patients in the second group were randomly and evenly divided into 10 subsets. We first established a fixed test set by randomly selecting two subsets. The remaining 8 subsets are used for cross-validation, with a training set to validation set ratio of 7:1. We selected 5 sets of training-validation sets for cross-validation. The images in second group were used for target task fine-tuning, and all metrics in this study were reported based on the fixed test set. The training and validation set of the second group and the first group were used for pre-training. As a result, the model never sees the test set in the pre-training stage, which avoids feature memorization or any form of information leakage. To comprehensively evaluate the proposed method, we further divided the training set into different proportions: 10\%, 20\%, and 50\%, containing 310, 621, and 1554 patients, respectively. We investigated the target performance with limited training images using different proportions of data.

\subsection{Compared methods}

We compared the proposed method with recent self-supervised methods on other images, including MoCo v2 \citep{he2020momentum}, C2L \citep{zhou2020comparing}, MG \citep{zhou2021models}, PCRL \citep{zhou2021preservational}, CAiD \citep{taher2022caid}, DiRA \citep{haghighi2022dira}, and SSFL \citep{li2020self}. As shown in Table.\ref{comm}, these methods were published in reputed journals or conferences and are current state-of-the-art (SOTA) self-supervised methods. These methods were re-implemented on our ultrasound images by referring to the authors' codes and articles. For a fair comparison, the original methods were uniformly modified while maintaining their specificity.

\subsection{Target tasks}

\begin{figure}[!t]
  \centering
  \includegraphics[scale=0.45]{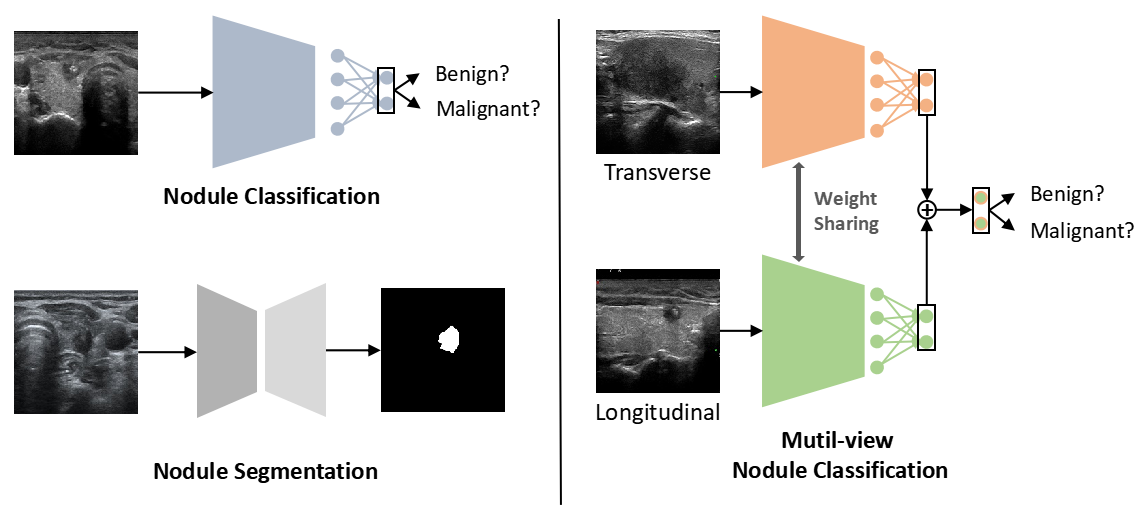}
  \caption{Networks of three target tasks. For NC, we use ResNet50 as the backbone and a one-layer fully connected layer as the classifier. For NS, we use the UNet as the network. For MNC, both two views have a network that consists of ResNet50 and a one-layer fully connected layer, and the two networks share the same weights. }
  \label{fig:target}
  \end{figure}

We employed three target tasks to evaluate the effectiveness of pre-training: nodule classification (NC), nodule segmentation (NS), and multi-view nodule classification (MNC). These networks are illustrated in Fig.\ref{fig:target}. For NC, we trained a model to predict benign and malignant nodules. We did not distinguish between the transverse and longitudinal images and treated them as independent samples. We used cross-entropy loss as the loss function. For NS, we adopted UNet \citep{ronneberger2015u} as the segmentation network. The encoder was ResNet50 and the decoder consisted of multiple convolutional layers and interpolation operations. We used Dice loss as the loss function. For MNC, we treated the transverse and longitudinal images of the same patient as the joint sample. The network consisted of two branches, corresponding to the two views. Each branch consists of ResNet50 and MLP. Both branches shared the same weights. This design suppresses overfitting by reducing the number of parameters, and its effectiveness has been proven \citep{wang2020auto, huang2022personalized}. We used the average of the outputs of the two branches as the final prediction results. We calculated the cross-entropy loss for the transverse branch, longitudinal branch, and final prediction, and we adopted the sum of the three losses as the final loss function.

We used the same data augmentation for the three target tasks: horizontal flip, brightness and contrast adjustment, random scale, rotation, and translation. The SGD optimizer and cosine learning rate decay scheduler were employed to optimize all models. We used an early stopping mechanism on the validation set to avoid overfitting. We set the weight decay to 0.00001 and the batch size to 32. For the randomly initialized models, we used the initial learning rate $\in \left \{0.01,0.02,0.05\right \}$ and training epochs $\in \left \{100,200\right \}$ to search for the best results as the baseline. For models using pre-trained weights, we used the initial learning rate $\in \left \{0.005,0.01\right \}$ and 100 epochs to train the models. We used the area under curve (AUC) score to evaluate nodule classification and the Dice score to evaluate nodule segmentation. Both pre-training and fine-tuning were performed five times.

\section{Results}

\begin{table}[!t]
  \caption{\label{nc} NC results (Unit:\%). The random model was trained from scratch with an initial learning rate of 0.02 and 200 training epochs. Other models were trained with an initial learning rate of 0.005.}
  \centering
  \begin{tabular}{p{1.8cm}p{1.1cm}p{1.1cm}p{1.1cm}p{1.1cm}}
  
  \toprule
  \textbf{Init} & \textbf{$\textit{r}$=10\%} & \textbf{$\textit{r}$=20\%} & \textbf{$\textit{r}$=50\%} & \textbf{$\textit{r}$=100\%}\\
  
  \midrule
  Random &
  60.28 &
  69.16 & 
  78.60 & 
  84.38 \\ 

  MoCo v2 &
  \underline{76.30} &
  \underline{78.40} & 
  82.78 & 
  85.52 \\ 
  
  C2L & 
  75.86 &
  78.21 & 
  82.84 & 
  86.01 \\ 
  
  MG & 
  54.86 &
  66.89 & 
  74.70 & 
  81.02 \\ 
  
  PCRL &
  74.02 &
  77.18 & 
  83.42 & 
  85.10 \\ 
  
  CAiD &
  74.30 &
  74.74 & 
  82.58 & 
  85.21 \\ 
  
  DiRA &
  75.87 & 
  78.19 & 
  \underline{83.59} & 
  \underline{86.28} \\ 
  
  SSFL & 
  75.95 &
  78.17 & 
  83.02 & 
  85.95 \\

  Ours & 
  $\textbf{77.75}^*$&
  $\textbf{79.87}^*$ & 
  $\textbf{84.36}^*$ & 
  \textbf{86.30} \\

  \bottomrule
  \end{tabular}
  \end{table}

  \begin{table}[!t]
    \caption{\label{ns} NS results (Unit:\%). The random model was trained from scratch with an initial learning rate of 0.02 and 100 training epochs. Other models were trained with an initial learning rate of 0.01. }
    \centering
  \begin{tabular}{p{1.8cm}p{1.1cm}p{1.1cm}p{1.1cm}p{1.1cm}}
    
  \toprule
  \textbf{Init} & \textbf{$\textit{r}$=10\%} & \textbf{$\textit{r}$=20\%} & \textbf{$\textit{r}$=50\%} & \textbf{$\textit{r}$=100\%}\\
        
    \midrule
    Random &
    68.11 &
    79.60 & 
    84.56 & 
    86.36  \\

    MoCo v2 &
    76.27 &
    80.58 & 
    \underline{84.72} & 
    \underline{86.46} \\ 
        
    C2L & 
    75.28 &
    80.42 & 
    84.17 & 
    86.01 \\ 
  
    MG & 
    66.99 &
    78.38 & 
    83.97 & 
    86.17 \\ 
  
    PCRL &
    75.63 &
    80.45 & 
    84.57 & 
    86.46 \\ 
        
    CAiD &
    73.08 &
    78.58 & 
    83.39 & 
    85.55 \\ 
  
    DiRA &
    75.06 &
    79.98 & 
    83.90 & 
    85.99 \\ 
        
    SSFL & 
    \underline{77.71} &
    \underline{81.76} & 
    84.66 & 
    86.44 \\ 
  
    Ours & 
    $\textbf{79.69}^*$ &
    $\textbf{82.31}^*$ & 
    \textbf{84.89} & 
    \textbf{86.50} \\ 
  
    \bottomrule
    \end{tabular}
    \end{table}

\begin{table}[!t]
      \caption{\label{mnc} MNC results (Unit:\%). The random model was trained from scratch with an initial learning rate of 0.01 and 200 training epochs. Other models were trained with an initial learning rate of 0.005. }
      \centering
  \begin{tabular}{p{1.8cm}p{1.1cm}p{1.1cm}p{1.1cm}p{1.1cm}}
      
  \toprule
  \textbf{Init} & \textbf{$\textit{r}$=10\%} & \textbf{$\textit{r}$=20\%} & \textbf{$\textit{r}$=50\%} & \textbf{$\textit{r}$=100\%}\\
    
    \midrule
      Random &
      58.90&
      66.80 & 
      78.37 & 
      86.69 \\ 
    
      MoCo v2 &
      77.28 &
      80.36 & 
      85.94 & 
      88.23 \\ 

      C2L & 
      \underline{77.43} &
      80.33 & 
      85.65 & 
      88.87 \\ 
        
      MG & 
      55.75 &
      62.43 & 
      72.32 & 
      81.69 \\ 
        
      PCRL &
      73.42 &
      78.87 & 
      85.60 & 
      88.30 \\ 
        
      CAiD &
      74.89 &
      78.76 & 
      83.46 & 
      87.69 \\ 
        
      DiRA &
      76.39 &
      78.27 & 
      \underline{86.70} & 
      \underline{89.33} \\ 
        
      SSFL & 
      74.41 &
      \underline{81.06} & 
      86.08 & 
      88.27 \\ 
    
      Ours & 
      $\textbf{78.17}^*$ &
      $\textbf{82.45}^*$ & 
      \textbf{87.08} & 
      \textbf{89.48} \\

      \bottomrule
      \end{tabular}
      \end{table}

In this section, we first compare the proposed method with SOTA methods and then compare the two-stage pre-training with ImageNet pre-training. For convenience, randomly initialized models are denoted as "Random". 
"SPIN" and "SSPIN" denote supervised and self-supervised pre-training on ImageNet, respectively. The two-stage pre-training is expressed in the form of “A$\to$B”. We use \textit{r} to represent the proportion of accessible data in the entire training set. Quantitative results are reported as the mean of five trials. The best and second-best results are bolded and underlined, respectively. Paired samples \textit{t}-test was performed.

\subsection{Nodule classification}

Table.\ref{nc} shows the results. The \textit{p}-values between our results and the second-best results were calculated. The symbol '*' indicates \textit{p}-value \textless 0.001 and is considered significant. The random model shows a large variance in performance when different proportions of the training data are used for training. Specifically, the random model achieved average AUC scores of 60.28\%, 69.16\%, 78.60\%, and 84.38\% with different proportions of training data (10\%, 20\%, 50\%, and 100\%), respectively. From 10\% to 100\%, the AUC score increased by 24.1\%. This shows that the random model is highly sensitive to the amount of training data, and insufficient data causes the model to perform poorly. In addition to the MG model, other self-supervised models have improved classification performance over the random model, and our model achieved the largest boost. Compared with the random model, our model improved the AUC scores by 17.47\%, 10.71\%, 5.76\%, and 1.92\%, respectively. Our model also significantly outperformed all other self-supervised models with limited manual labels, demonstrating its effectiveness.

\begin{figure*}[!t]
  \centering
  \includegraphics[scale=0.36]{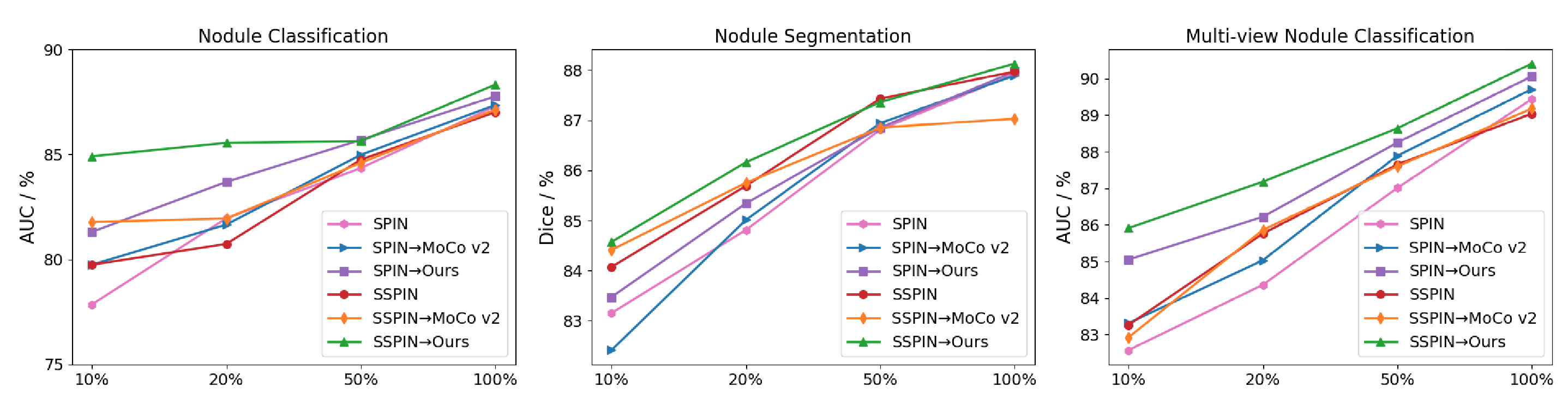}
  \caption{Comparison between ImageNet pre-training and two-stage pre-training, based on three target tasks and different proportions of training data.}
  \label{fig:line_total}
\end{figure*}

\subsection{Nodule segmentation}

Table.\ref{ns} presents the nodule segmentation results. The performance of the random model varies significantly for different proportions of training data. This indicates that the random model performs poorly on a few training datasets. In contrast, using less training data, self-supervised models other than the MG model improve segmentation performance. Our model comprehensively outperforms other self-supervised models and achieves average Dice scores of 79.69\%, 82.31\%, 84.89\%, and 86.50\%. This shows that our method can also significantly improve nodule segmentation with limited manual labels, demonstrating its effectiveness. 

\subsection{Multi-view nodule classification}

Table.\ref{mnc} presents the multi-view nodule classification results. Similarly, for different proportions of training data, the random model exhibits a large variance in performance. The score difference between 10\% and 100\% of the training data is 27.79\%. Compared with the random model, the self-supervised models improve the multi-view nodule classification performance, except for the MG model. The DiRA model, for example, improves the AUC scores by 17.49\%, 11.47\%, 8.33\%, and 2.64\%. This is a major improvement, particularly with less training data. Our model achieves average AUC scores of 78.17\%, 82.45\%, 87.08\%, and 89.48\%, which significantly outperforms other self-supervised models when using a small amount of training data (i.e. \textit{r}=10\% and 20\%). This demonstrates the effectiveness of the proposed method.

\subsection{Two-stage pre-training}

\begin{table}[!t]
  \caption{\label{nc2} NC (two-stage) results (Unit:\%). All models were trained with an initial learning rate of 0.005.}
  \centering
  \begin{tabular}{p{2.4cm}p{1.0cm}p{1.0cm}p{1.0cm}p{1.0cm}}

  \toprule
  \textbf{Init} & \textbf{$\textit{r}$=10\%} & \textbf{$\textit{r}$=20\%} & \textbf{$\textit{r}$=50\%} & \textbf{$\textit{r}$=100\%}\\

 \midrule
  SPIN & 
  77.86 &
  81.96 & 
  84.35 & 
  87.29 \\ 

  SPIN$\to$MoCo v2  &
  79.75 &
  81.66 &
  84.98 &
  87.35 \\

  SPIN$\to$Ours & 
  $\textbf{81.31}^\dagger$ &
  $\textbf{83.69}^\dagger$ & 
  $\textbf{85.69}^\dagger$ & 
  $\textbf{87.76}$ \\ 

  \midrule
  SSPIN & 
  79.75  &
  80.74  & 
  84.75  & 
  87.01  \\ 

  SSPIN$\to$MoCo v2 &
  81.78  &
  81.95  & 
  84.60  & 
  87.13  \\

  SSPIN$\to$Ours & 
  $\textbf{84.91}^\dagger$ &
  $\textbf{85.55}^\dagger$ & 
  $\textbf{85.63}^\dagger$ & 
  $\textbf{88.32}^\dagger$ \\ 

  \bottomrule
  \end{tabular}
  \end{table}

\begin{table}[htpb]
  \caption{\label{ns2} NS (two-stage) results (Unit:\%). All models were trained with an initial learning rate of 0.01.}
  \centering
  \begin{tabular}{p{2.4cm}p{1.0cm}p{1.0cm}p{1.0cm}p{1.0cm}}

  \toprule
  \textbf{Init} & \textbf{$\textit{r}$=10\%} & \textbf{$\textit{r}$=20\%} & \textbf{$\textit{r}$=50\%} & \textbf{$\textit{r}$=100\%}\\

  \midrule

  SPIN &
  83.15 &
  84.81 & 
  86.81 & 
  87.92 \\

  SPIN$\to$MoCo v2  &
  82.41 &
  85.01 & 
  \textbf{86.94} & 
  87.89 \\ 

    SPIN$\to$Ours & 
  $\textbf{83.47}$ &
  $\textbf{85.34}^\dagger$ & 
  86.85 & 
  $\textbf{87.97}$ \\ 

  \midrule
  SSPIN & 
  84.07 &
  85.69 & 
  \textbf{87.43} & 
  87.97 \\ 

  SSPIN$\to$MoCo v2 & 
  84.41 &
  85.75 & 
  86.85 & 
  87.03\\

  SSPIN$\to$Ours &
  % $\mathrm{Ours}_{\mathrm{selfsupIN}}$ & 
  $\textbf{84.57}^\dagger$ &
  $\textbf{86.16}^\dagger$ & 
  87.36 & 
  $\textbf{88.13}$ \\ 

  \bottomrule
  \end{tabular}
  \end{table}

The nodule classification, segmentation, and multi-view nodule classification results are presented in Tables.\ref{nc2}, \ref{ns2}, and \ref{mnc2}, respectively. The \textit{p}-values between two-stage pre-training of our method and ImageNet pre-training were calculated. The symbol '$\dagger$' indicates \textit{p}-value \textless 0.001 and is considered significant. To present the results better, we drew line charts for the three tasks, as shown in Fig.\ref{fig:line_total}. Compared to ImageNet pre-training, the two-stage pre-training of our method almost always improves performance. For example, the AUC scores of SPIN$\to$Ours are 3.45\%, 1.73\%, 1.34\%, and 0.47\% higher than SPIN in NC. In NS, SSPIN$\to$Ours is only slightly lower than SSPIN when \textit{r}=50\%, and outperforms SSPIN in other proportions. In MNC, SPIN$\to$Ours improved the AUC scores by 2.48\%, 1.86\%, 1.24\%, and 0.62\%, respectively. Compared to ImageNet pre-training, MoCo v2's two-stage pre-training improves performance or reaches competitive performance. Two-stage pre-training of our method outperforms that of MoCo v2 in most cases, demonstrating the effectiveness of the proposed method.

\begin{table}[!t]
  \caption{\label{mnc2} MNC (two-stage) results (Unit:\%). All models were trained with an initial learning rate of 0.005.}
  \centering
  \begin{tabular}{p{2.4cm}p{1.0cm}p{1.0cm}p{1.0cm}p{1.0cm}}
    
  \toprule
  \textbf{Init} & \textbf{$\textit{r}$=10\%} & \textbf{$\textit{r}$=20\%} & \textbf{$\textit{r}$=50\%} & \textbf{$\textit{r}$=100\%}\\
    
  \midrule

  SPIN &
  82.57 &
  84.36 & 
  87.01 & 
  89.44 \\ 

  SPIN$\to$MoCo v2  &
  83.31 &
  85.03 & 
  87.89 & 
  89.71 \\

  SPIN$\to$Ours & 
  % $\mathrm{Ours}_{\mathrm{supIN}}$ & 
  $\textbf{85.05}^\dagger$ &
  $\textbf{86.22}^\dagger$ & 
  $\textbf{88.25}^\dagger$& 
  $\textbf{90.06}^\dagger$\\

  \midrule
  SSPIN & 
  83.26 &
  85.76 & 
  87.65 & 
  89.04 \\ 

  SSPIN$\to$MoCo v2 &
  82.92 &
  85.86 & 
  87.60 & 
  89.18 \\

  SSPIN$\to$Ours & 
  % $\mathrm{Ours}_{\mathrm{selfsupIN}}$ & 
  $\textbf{85.91}^\dagger$&
  $\textbf{87.18}^\dagger$&
  $\textbf{88.64}^\dagger$&
  $\textbf{90.41}^\dagger$\\ 
    
  \bottomrule
  \end{tabular}
  \end{table}

\subsection{Ablation study}

Table.\ref{lam} lists the results of three downstream tasks with different lambda values in our method. When lambda is equal to 0, the method degenerates to MoCo v2. This model performs worst in given lambda values. This demonstrates the effectiveness of cross-view contrastive learning. Our method achieves the best performance when lambda is equal to 0.5. We also verify the impact of the number of paired views on cross-view contrastive loss. The results are shown in Table.\ref{miss}. From the table we can observe: (1) Our method always outperforms MoCo v2 even though MoCo v2 uses all unpaired views while our method does not use any unpaired views. This shows that cross-view contrastive loss is important. (2) Our approach continues to improve performance as available unpaired views increase. This demonstrates the effectiveness of the proposed adaptive loss function. Considering the clinical practice of coexistence of multiple views and missing views in thyroid ultrasound examination, our method can help improve thyroid ultrasound diagnosis with limited labeled data.

  \begin{figure*}[!t]
    \centering
    \includegraphics[scale=0.55]{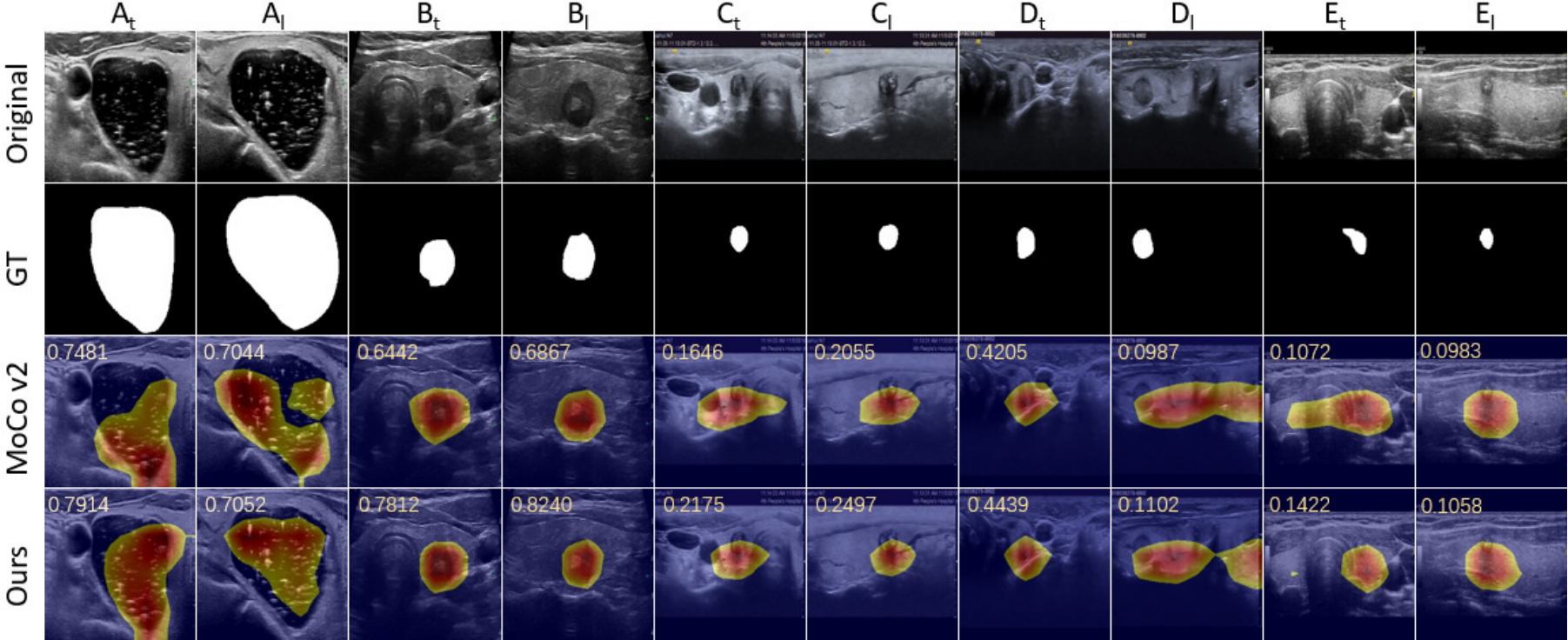}
    \caption{Activation map visualization. The first row is the original image, and the second row is the corresponding mask. The third and fourth rows are the activation maps of MoCo v2 and our method, respectively. Numbers on the activation map represent Dice scores. We show five pairs of images and use a threshold of 0.5.}
    \label{fig:camseg}
    \end{figure*}

\begin{table}[h]
  \caption{\label{lam} Results (Unit:\%) of three downstream tasks with different lambda values in our method.}
  \centering
  \begin{tabular}{p{1.0cm}p{1.0cm}p{1.0cm}p{1.0cm}p{1.0cm}p{1.0cm}}

  \toprule
  \multirow{2}{*}{\textbf{Tasks}}& 
  \multirow{2}{*}{\textbf{\textit{r}}}& 
  \multicolumn{4}{c}{$\textbf{$\lambda$ in Equation 6}$} \\ 
  \cline{3-6} &&
  \textbf{0}   & 
  \textbf{0.2} & 
  \textbf{0.5} & 
  \textbf{1.0} \\

 \midrule

  \multirow{2}{*}{NC} &
   10\% &
  76.30 &
  77.67& 
  \textbf{77.75}& 
 76.79\\

    &
   20\% &
  78.40 &
  79.71& 
  \textbf{79.87}& 
  79.23\\ 

 \midrule
    \multirow{2}{*}{NS} &
     10\% &
    76.27 &
    78.89 & 
    \textbf{79.69} & 
     79.56\\ 
    
     &
    20\% &
     80.58&
     81.97& 
     82.31& 
     \textbf{82.43}\\ 

 \midrule
  \multirow{2}{*}{MNC} &
   10\% &
  77.28&
  77.52& 
  \textbf{78.17}& 
  77.47 \\ 

   &
   20\% &
  80.36&
  81.79& 
  \textbf{82.45}& 
  82.29\\ 

  \bottomrule
  \end{tabular}
  \end{table}

\begin{table}[!h]
  \caption{\label{miss}  NC results (Unit:\%). The dataset used for pre-training consists of all paired images (training and validation set in the second group) and different proportions of 779 unpaired views.}
  \centering
  \begin{tabular}{p{0.8cm}p{1.5cm}p{0.9cm}p{0.9cm}p{0.9cm}p{0.9cm}}

  \toprule
  \multirow{2}{*}{\textbf{\textit{r}}}& 
  \multirow{2}{*}{\textbf{Init}}& 
  \multicolumn{4}{c}{$\textbf{+ (\%) of 779 unpaired views}$} \\ 
  \cline{3-6} &&
  \textbf{0}   & 
  \textbf{20\%} & 
  \textbf{50\%} & 
  \textbf{100\%} \\

 \midrule

  \multirow{2}{*}{10\%} &
   MoCo v2 &
   75.02&
  75.14& 
  75.67& 
 76.30\\

    &
   Ours &
  $77.13$ &
  $77.20$ &
  $77.54$ &
  $77.75$ \\ 

 \midrule
    \multirow{2}{*}{20\%} &
     MoCo v2 &
     76.08&
     76.46& 
     78.41& 
     78.40\\ 
    
     &
    Ours &
    $79.33$ &
    $79.23$ &
    $79.79$ &
    $79.87$ \\ 

 \midrule
    \multirow{2}{*}{50\%} &
     MoCo v2 &
     81.34&
     81.83& 
     81.94& 
     82.78\\ 
    
     &
    Ours &
    $83.76$ &
    $84.04$ &
    $84.29$ &
    $84.36$ \\ 

 \midrule
    \multirow{2}{*}{10\%} &
     MoCo v2 &
     85.37&
     85.33& 
     85.45& 
     85.52\\ 
    
     &
    Ours &
    $86.14$ &
     $86.34$ &
     $86.30$ &
     $86.50$ \\ 

  \bottomrule
  \end{tabular}
  \end{table}

\section{Discussions}

\subsection{Why is our method better?}

Our method significantly outperformed those designed for single-view images. This could be because our pre-training method makes the model pay more attention to the nodule area, which provides a good prior for the three target tasks. To verify this, we used activation maps as nodules' segmentation maps and computed the Dice score using nodule masks. Specifically, we froze the pre-trained ResNet50 and fed all images from the test set into it. We obtained the feature maps before the global pooling layer. The activation map is obtained by directly averaging the feature maps along the channel dimension. We resized the activation map to the original image size and normalized it to [0,1]. We obtained the segmentation map by binarizing the activation map with different thresholds $\textit{t}\in \left \{ 0.3,0.4,0.5,0.6,0.7 \right\}$. We compared our method with MoCo v2. The quantitative results are presented in Table.\ref{camdice} and the qualitative visualizations are shown in Fig.\ref{fig:camseg}. Our method always has higher Dice scores than MoCo v2 and pays more attention to the nodule area, although nodule sizes vary significantly. Several studies have indicated that lesion segmentation facilitates accurate disease classification \citep{zhang2021bi, zhou2021multi}. This may explain why our method achieves better classification and segmentation performance. In addition, our method also outperforms SSFL \citep{li2020self}, which also adopts multi-view contrastive learning. This is because SSFL can only utilize paired data, and images with only one transverse or one longitudinal view are not utilized. Overall, our method benefits from multi-view contrastive learning that eliminates the paired data constraints.

\begin{table}[!t]
  \caption{\label{camdice} Average Dice score between the activation maps and nodule masks at different thresholds (Unit:\%).}
  \centering
  \begin{tabular}{p{1.9cm}p{0.8cm}p{0.8cm}p{0.8cm}p{0.8cm}p{0.8cm}}

  \toprule
  \textbf{Pre-training} & \textbf{$\textit{t}$=0.3} & \textbf{$\textit{t}$=0.4} & \textbf{$\textit{t}$=0.5} & \textbf{$\textit{t}$=0.6} & \textbf{$\textit{t}$=0.7}\\

  \midrule
  MoCo v2 & 
  32.01 &
  34.61 & 
  35.93 & 
  35.86 &
  33.75 \\ 

  Ours & 
  \textbf{36.42} &
  \textbf{39.50} & 
  \textbf{41.11} & 
  \textbf{40.96} &
  \textbf{38.20} \\ 

  \bottomrule
  \end{tabular}
  \end{table}

\begin{figure*}[h]
    \centering
    \includegraphics[scale=0.50]{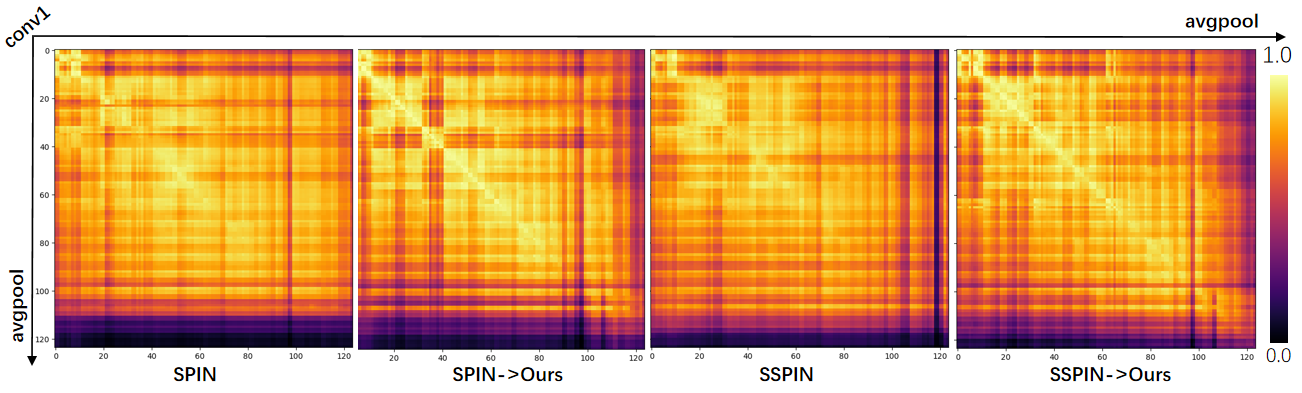}
    \caption{Centered kernel alignment (CKA) score map. Based on nodule classification, we calculate the CKA scores of the different layers of models before and after fine-tuning, including convolutional, batchnorm, and pooling layers. The value at coordinates (\textit{i}, \textit{j}) in each map represents the CKA score between the \textit{i}-th layer of the model before fine-tuning the \textit{j}-th layer of the model after fine-tuning. }
    \label{fig:cka}
\end{figure*}

\begin{table}[t]
  \caption{\label{cka} Comparison of feature reuse between two-stage pre-training and ImageNet pre-training. Each row presents the CKA score for different intermediate layers before and after fine-tuning models in nodule classification.}
  \centering
  \begin{tabular}{p{1.9cm}p{0.8cm}p{0.8cm}p{0.8cm}p{0.8cm}p{0.8cm}}

  \toprule
  \textbf{Pre-training} & \textbf{conv1} & \textbf{layer1} & \textbf{layer2} & \textbf{layer3} & \textbf{layer4}\\

 \midrule
  SPIN & 
  0.966&
  0.920& 
  0.965& 
  \textbf{0.911}& 
  0.114 \\ 

  %\hline
  SPIN$\to$Ours & 
  \textbf{0.993}&
  \textbf{0.993}& 
  \textbf{0.991}& 
  0.846& 
  \textbf{0.179}\\ 

  \midrule
  SSPIN & 
  0.942&
  0.947& 
  0.957& 
  0.854& 
  0.126 \\ 

  %\hline
  SSPIN$\to$Ours & 
  \textbf{0.975}&
  \textbf{0.989}& 
  \textbf{0.982}& 
  \textbf{0.937}& 
  \textbf{0.381} \\ 

  \bottomrule
  \end{tabular}
  \end{table}

\subsection{Why is two-stage pre-training better?}

The two-stage pre-training uses ImageNet pre-training and significantly surpasses it without using additional labels. Good pre-trained weights provide more reusable features \citep{neyshabur2020being}. Following \citep{neyshabur2020being}, we evaluated the degree of feature reuse by measuring the feature similarity of the different layers of the models before and after fine-tuning using centered kernel alignment (CKA) \citep{kornblith2019similarity}. Fig.\ref{fig:cka} shows the visualization results. A higher CKA score indicates more feature reuse, and we primarily focused on the CKA scores at the diagonal positions. Two-stage models have higher CKA scores than corresponding ImageNet pre-training models, especially in the low/mid-layer. This shows that the two-stage pre-training provides more reusable features than the ImageNet pre-training. We further computed the CKA scores of several representative layers, as shown in Table.\ref{cka}. The two-stage model corresponding to SSPIN improves the CKA scores across the board more significantly in the highest layer. It is well known that in CNNs, the lower layers extract detailed features, while the higher layers extract task-specific features. This shows that the two-stage self-supervised pre-training not only provides more reusable detailed features but also provides more reusable task-specific features. This is probably why it performs better than ImageNet pre-training, even without using any additional labels.

\section{Conclusion}

We proposed a multi-view contrastive self-supervised method to improve the nodule classification and segmentation performance of thyroid ultrasound images with limited manual labels. Our method enables the model to learn transformation- and view-invariant features. To address the issue of missing views, we designed an adaptive loss function that eliminates the need for paired views. We also adopted a two-stage pre-training strategy to alleviate the domain shift between natural and medical images. To verify the effectiveness of the proposed method, we constructed a large-scale thyroid ultrasound image dataset from more than 20 hospitals. The results of the extensive experiments show that the proposed method significantly improves nodule classification and segmentation performance compared to random initialization and outperforms other SOTA self-supervised methods with limited manual labels. The results also show that the two-stage pre-training strategy can significantly boost the target performance.

\section*{Conflict of interest statement}
The authors declare that they have no known competing financial interests or personal relationships that could have appeared to influence the work reported in this paper.

\section*{Acknowledgments}
This work was supported by the National Natural Science Innovative Research Group Project (61821002), the Key Project of the National Natural Science Foundation of China (51832001), and the Frontier Fundamental Research Program of Jiangsu Province for Leading Technology (BK20222002).

% Use the style of numbering in square brackets.
% If nothing is used, default style will be taken.
%\begin{enumerate}[a)]
%\item 
%\item 
%\item 
%\end{enumerate}  

%\begin{itemize}
%\item 
%\item 
%\item 
%\end{itemize}  

%\begin{description}
%\item[]
%\item[] 
%\item[] 
%\end{description}  

% Figure
% \begin{figure}[<options>]
% 	\centering
% 		\includegraphics[<options>]{}
% 	  \caption{}\label{fig1}
% \end{figure}

% \begin{table}[<options>]
% \begin{tabular*}{\tblwidth}{@{}LL@{}}
% \toprule
% \midrule
%  & \\
%  & \\
%  & \\
%  & \\
% \bottomrule
% \end{tabular*}
% \end{table}

% Uncomment and use as the case may be
%\begin{theorem} 
%\end{theorem}

% Uncomment and use as the case may be
%\begin{lemma} 
%\end{lemma}


% \section{}\label{}


\bibliographystyle{cas-model2-names}

\bibliography{cas-refs}

\bio{}
\endbio

\endbio

\end{document}